\pdfoutput=1 % ensure pdflatex (for arXiv)

\documentclass[11pt]{article}
\usepackage[]{acl}
\usepackage{nert}
\usepackage{times}
\usepackage{latexsym}
\usepackage{amssymb}% http://ctan.org/pkg/amssymb
\usepackage{pifont}% http://ctan.org/pkg/pifont
\usepackage[inline,shortlabels]{enumitem}
\usepackage{comment}
\usepackage{multirow}

\usepackage{todonotes}
\usepackage{microtype}
\usepackage{xcolor}

\usepackage{kotex}

\newcounter{mycounter}

\title{SPT: Semi-Parametric Prompt Tuning for Multitask Prompted Learning}
\newcommand*\samethanks[1][\value{footnote}]{\footnotemark[#1]}

\author{
\textbf{M Saiful Bari}\Thanks{Work done at Amazon Web Services.}$^{\ \P}$\Thanks{Corresponding authors: bari0001@e.ntu.edu.sg, astonz@amazon.com } , 
\textbf{Aston Zhang}$^{\S}$\samethanks \ , 
\textbf{Shuai Zheng}$^\S$,  
\textbf{Xingjian Shi}$^\S$,
\\
\textbf{Yi Zhu}$^\S$, 
\textbf{Shafiq Joty}$^{\P}$, 
\textbf{Mu Li}$^\S$, 
\\
$^{\P}$Nanyang Technological University \;
$^\S$Amazon Web Services \;
}

\begin{document}
\maketitle
\begin{abstract}
Pre-trained large language models can efficiently interpolate human-written prompts in a natural way.
Multitask prompted learning can help generalization through a diverse set of tasks at once, thus enhancing the potential for more effective downstream fine-tuning. To perform efficient multitask-inference in the same batch, parameter-efficient fine-tuning methods such as prompt tuning have been proposed. However, the existing prompt tuning methods may lack
generalization.
We propose SPT, a semi-parametric prompt tuning method for multitask prompted learning. 
The novel component of SPT is a memory bank from where memory prompts are retrieved based on discrete prompts.
Extensive experiments,
such as (i) fine-tuning a full language model with SPT on 31 different tasks from 8 different domains and evaluating zero-shot generalization on 9 heldout datasets under 5 NLP task categories and (ii) pretraining SPT on the GLUE datasets and evaluating fine-tuning on the SuperGLUE datasets, 
demonstrate effectiveness of SPT.
\end{abstract}
\section{Introduction}

Large language models (LLMs) have shown emergent capabilities that are solely learned from raw texts \cite{gpt-3,kim_b,emergent_jason_wei}.
Upon performing pre-training with self-supervised objectives \cite{wang_lm_obj,ul2,glm,alexatm20b}, LLMs can efficiently interpolate human-written task instructions or prompts in a natural way. At scale, these well-written prompts encourage an LLM to generate relevant texts or even perform complex reasoning with proper contexts \cite{CoT,zhang2022automatic}. However, such prompts often require a few examples or demonstrations to exhibit emergent behavior.

\begin{figure*}
    % https://drive.google.com/file/d/1vbn8KW5TU-rmHwgkFAHIhVxhF7H-a8lu/view?usp=sharing
    \includegraphics[scale=.9]{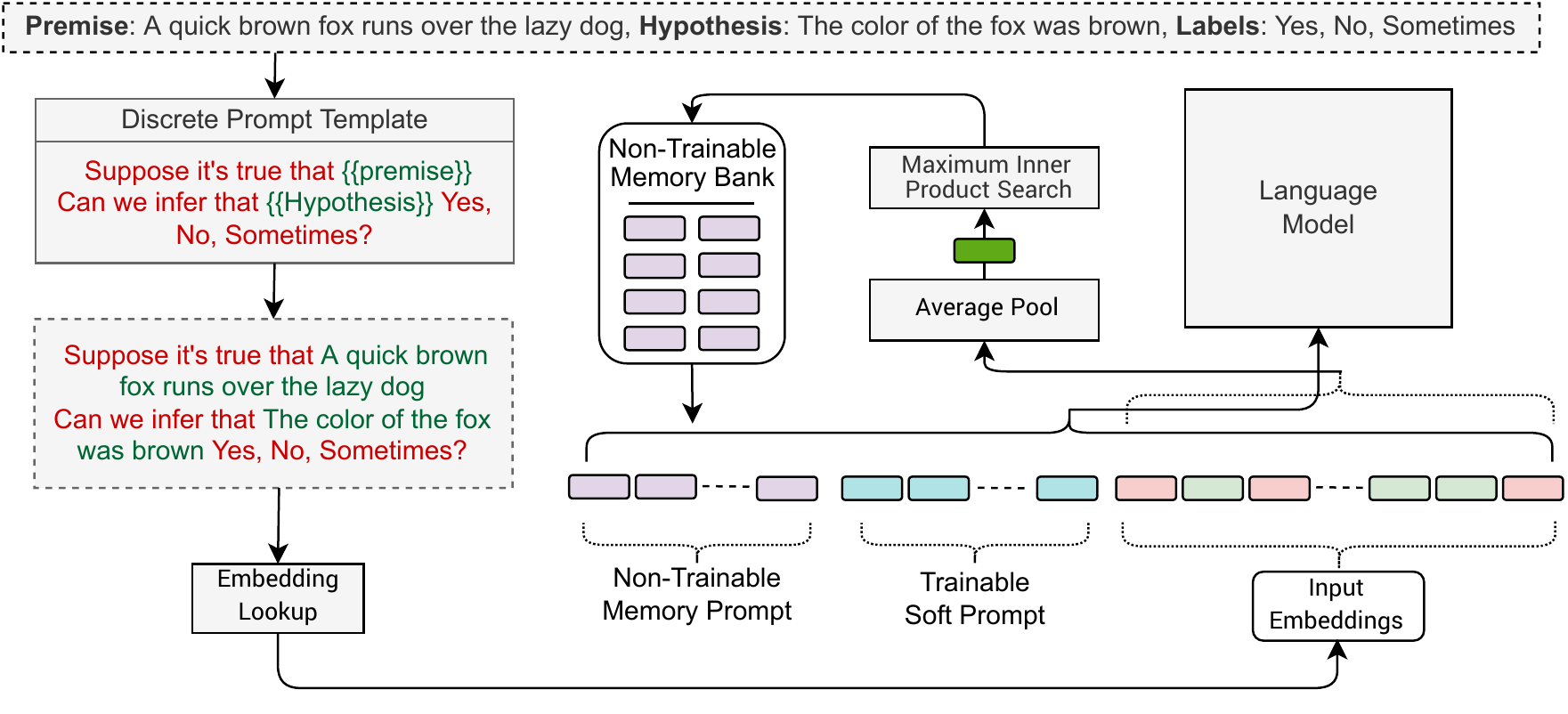}
    \caption{Overview of the SPT (semi-parametric prompt tuning) method. The memory bank is a frozen deep copy of the language model's embedding layer, where the aggregated discrete prompt information is used to retrieve the most relevant token embeddings to form the memory prompt. The memory prompt, soft prompt, and discrete prompt (input embeddings) are concatenated as the input to the language model. For multitask-inference full fine-tuning, we do not use the soft-prompt. For multitask-inference parameter-efficient fine-tuning, soft-prompts are meta-trained and finetuned.}
    \label{fig:model}
\end{figure*}

One way to address this issue is to apply \textbf{multitask prompted learning}, where inputs to a set of multiple tasks are transformed through expertly written prompt templates \cite{promptsource,super_natural}, and then fine-tune the LLM on those instantiated prompts \cite{FLAN,T0,Flan-T5,bloomz}. 
For example, a \textbf{discrete prompt template}
for a natural language inference (NLI) task (\Cref{fig:model})
can be: ``\emph{Suppose it's true that} \{\{\texttt{premise}\}\} \emph{Can we infer that} \{\{\texttt{hypothesis}\}\} \emph{Yes, No, Sometimes?}''.
By inserting a hypothesis and a premise from 
an NLI task instance,
we obtain an instantiated \textbf{prompt},
such as
``\emph{Suppose it's true that A quick brown
fox runs over the lazy dog
Can we infer that The color of the fox
was brown Yes, No, Sometimes?}'' In multitask prompted learning, we use such instantiated prompts for multiple tasks to finetune a LLM. This meta-training often helps LLMs generalize well to unseen tasks but still lacks full downstream task finetuning performance \cite{liu2020tfew}.

%To perform MTPL, we transform a task inputs with  

\begin{figure}
    \centering
    % https://drive.google.com/file/d/1zS3q829hriGdSEadB79fOpZj7dvUdgZG/view?usp=sharing
    \includegraphics[scale=.45]{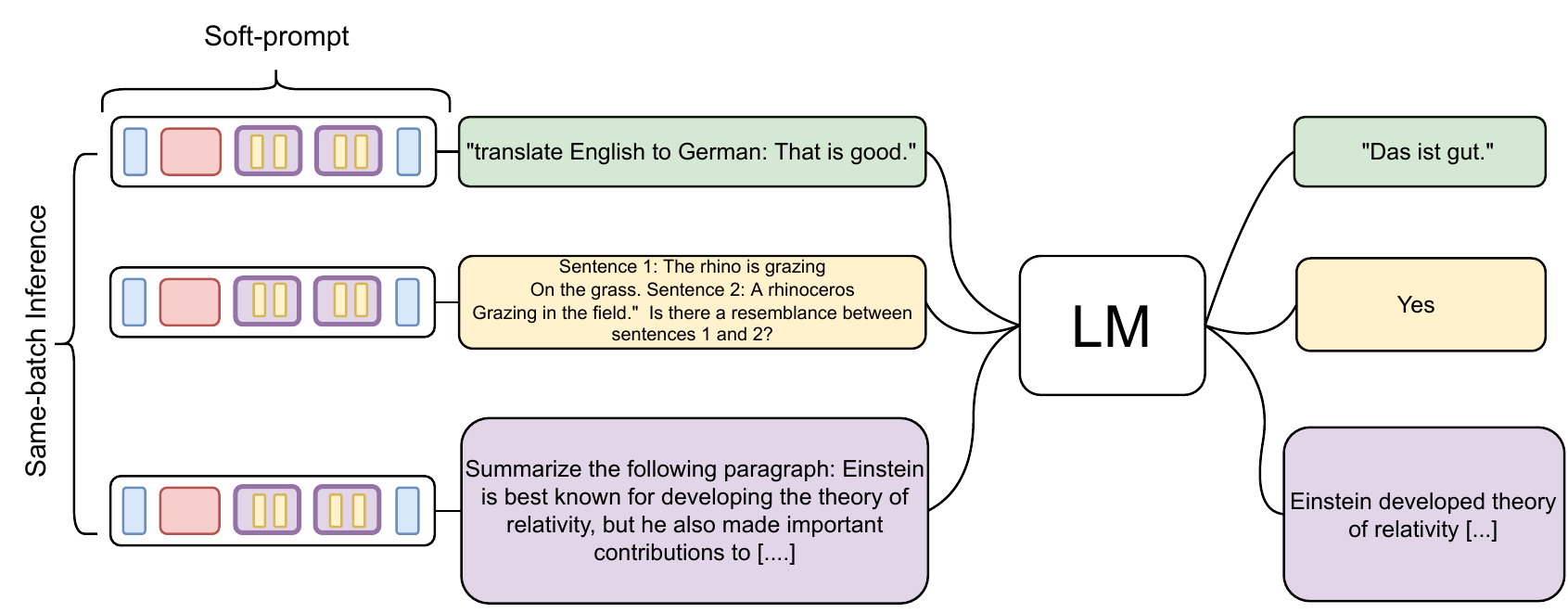}
    \caption{An example of same-batch multitask-inference. For each task sample in the same batch, a soft prompt is prepended to the task input. The same frozen language model (LM) is conditioned on each of the task-specific soft prompts separately, which specifies inference for each task in the batch. }
    \label{fig:pemi}
\end{figure}

On the other hand, from the perspective of efficiency and feasibility, as the model sizes grow like GPT-3, updating all the parameters of an LLM may not be a feasible option. This applies to both downstream and upstream (i.e., meta-training) finetuning scenarios. For downstream task finetuning, full model finetuning may also run into the risk of overfitting  \cite{pfeiffer2020madx}. To address this, a number of parameter-efficient finetuning methods have been proposed. One class of methods add extra  tunable layers \cite{li2021prefixtuning,adapter_fusion} and/or tune only a few set of selected parameters  \cite{adapter_hub,le2021parameterized,fishmask,hu2021lora,bitfit}. However, these approaches 
have one major limitation:
as the internal layers become task specific, they do not allow multitask-inference in the same batch (Figure \ref{fig:pemi}). For performing inference with LLMs on multiple tasks at an accelerated device, these methods become inefficient in practical settings since they require model loading/unloading (when switching tasks) from the device and performing only single task inference in a batch.

%Many NLP tasks have achieved state-of-the-art results by finetuning pre-trained language models \cite{wang_glue,wang2020superglue,hu2020xtreme,Liang2020XGLUEAN}. 

%However, as the models grow larger, updating all the parameters for a small task becomes computationally expensive and inefficient. 

%To combat this problem during training, many Parameter Efficient Finetuning (PEFT) methods have been proposed recently \cite{pfeiffer2020madx,le2021parameterized,fishmask,hu2021lora}, updating or adding only a few set of specific manually identified trainable weights. 

%However, these models still have a performance issue during inference. 

%In contrast to PEFT methods, we refer to parameter-efficient methods that can perform multitask-inference in the same batch as the Parameter Efficient multitask-inference (PEMI) methods. Figure \ref{fig:pemi} shows the basic structure of a PEMI based method.

Another class of parameter-efficient finetuning methods is \textbf{prompt tuning} or PT \cite{shin2020autoprompt,schick-schutze-2021-just,gao-prompt,WARP}, originally proposed for small-scale language models (LMs), and later scaled to LLMs by \citet{lester_prompt}. In this approach, the language model is kept frozen, and only the sequence of prompt token embeddings (a.k.a., \textbf{soft prompt}) that are prepended to the input embeddings are tuned.  Figure \ref{fig:pemi} shows an example. Since the language model is kept frozen in the finetuning process, it supports efficient same-batch multitask-inference and can adapt well to downstream requirements while preserving the language model's generic distribution \cite{qin2022lfpt}. However, its performance has been shown to be susceptible to the initialization of prompt embeddings \cite{prompt_pretrain}.

In view of this,
\citet{SpoT} and \citet{asai2022attempt}  propose methods to transfer prompts from a set of pre-learned prompt embedding templates to initialize a soft prompt for a new target task. 
%Both of these parameter-efficient finetuning methods freeze all of the LM's parameters, allowing them to perform inference for multiple tasks in the same batch. As such, parameter-efficient finetuning methods that can perform multitask-inference in the same batch are also called \textbf{multitask-inference parameter-efficient finetuning} methods.
To the best of our knowledge, the primary objective of employing the PT methods has so far been to train a soft prompt only on a particular downstream task. We hypothesize that similar to full-model finetuning \cite{T0,FLAN}, the meta-training in the form of multi-task prompted learning can also benefit parameter-efficient PT methods to achieve better generalization. However, one impediment to achieve this is the strict structural requirement, i.e., the fact that it must be a sequence of soft tokens, which may limit the prompt's expressiveness. Furthermore, the templatized nature of the discrete prompts may induce bias in the prompt space \cite{albert_prompt_usecase}.

%This memory bank is introduced to mitigate the issue that a strictly structured discrete prompt template may induce bias and lack generalization in the prompt space \cite{albert_prompt_usecase}. 

%The structured form  (i.e., strictly specific sequential tokens) of a soft prompt may generalize well over a small target task but may lack generalization in 
%multitask-inference finetuning \cite{SpoT}.

% \shafiq{This and the following two paragraphs need to be revised to make them coherent.} -> \maruf{re-ordering the passages for making it coherent}

% \shafiq{How is SPT connected to PEFT? If it is not directly connected, you can simply talk about PT and its PEMI capability, and move the whole PEFT to Rel Work.} ->\maruf{Please follow the re-ordering}

%\shafiq{The motivation for SPT is missing: why we need semi-parametric? } -> \maruf{added}

% \shafiq{what are the limitations (e.g., not being instance-specific?) of PT that SPT can address to get better performance?} \maruf{ added}

%\shafiq{ Are there similar methods ( I know a few) that try to address the same issue? Why you think your method is superior?} maruf{Not sure what you meant? Are you referring to Automatic prompt selection?}

In this work, we propose \textbf{SPT}: semi-parametric PT for multitask prompted learning. 
The novel component of SPT is a \emph{non-trainable} memory bank (thus ``semi-parametric'' in the name), 
from where memory prompts are retrieved based on the encoding of prompted inputs.
%This memory bank is introduced to mitigate the issue that a strictly structured discrete prompt template may induce bias and lack generalization in the prompt space \cite{albert_prompt_usecase}. 
More precisely, we initialize the memory bank with the model's embedding layer and kept it frozen all the time.  Based on the embeddings of the prompted (discrete) input, we perform  maximum inner product search to retrieve most similar tokens from the memory as an input-dependent memory prompt (\Cref{fig:model}).

%We use the average representation of the discretely prompted input samples and use them to run a maximum inner product search to retrieve top $k_1$ tokens from the memory bank.

The key idea behind our design is that a way to improve the performance of an LM is to give it access to additional context (from the memory) that can help it make more informed predictions. The sparse semi-parametric nature of the memory prompts enables the LM to cover robust prompt distribution with the help of discrete prompts whose embeddings are trained with supervision. 

%Thus, instead of doing PT of an LM on a single target task, we employ PT in a multitask prompted meta-training stage. 
 %about the next token in a sequence

We evaluate SPT under two challenging experimental settings:
(i) \textbf{multitask-inference full finetuning}:
we fine-tune a full LM, especifically T5 \cite{2020t5}, with SPT on the T0 split \cite{T0} of the P3 dataset consisting of 31 different tasks from 8 different domains,
then evaluate its zero-shot generalization performance on 9 held-out datasets under 5 task categories: NLI, coreference resolution, word sense disambiguation, sentence completion, and question answering;
and
(ii) \textbf{multitask-inference parameter-efficient fine-tuning}:
we meta-train the soft prompts with SPT on the GLUE datasets \cite{wang_glue},
then fine-tune them on the SuperGLUE datasets \cite{wang2020superglue}.
With extensive experiments and empirical analysis, we demonstrate the effectiveness of the proposed SPT method. 

All in all, we make the following contributions:

\begin{itemize}
    \item We propose SPT, a semi-parametric prompt tuning method for multitask prompted learning. SPT uses a memory bank to retrieve relevant memory prompts based on the embeddings of the discrete prompts.
    \item We fine-tune a full LM with SPT on 31 different tasks from 8 different domains and evaluate zero-shot generalization on 9 heldout datasets under 5 NLP task categories. We also meta-train soft-prompts with SPT on the GLUE datasets and evaluate their finetuning performance on the SuperGLUE datasets. 
    \item We find that the distributions of SPT prompts are well spread and not clustered based on any tasks, which indicates a successful multitask pre-training. 
    %\shafiq{give main findings..}
\end{itemize}

%Unlike prompt tuning, where prompt templates are task-dependent, we propose task-invariant prompting, which helps ground the multi-task samples into a single task in a more generalized way. The retrieved top-$k$ embeddings are prepended with the regular inputs of the language model. These prepended memory tokens help the language model perform better inference for unseen tasks.

% Let $s$ and $t$ denote the source and target language distributions, respectively. 

% \shafiq{s and t are never used except in the Eq, which I believe you mean train} - > Addressed

% Let $\mathcal{D}_{train} = \{(x_i^t, y_i^t)\}_{i=1}^{N}$ and $\mathcal{D}_{test} = \{(x_i^t\}_{i=1}^{M}$ refer to the training and test (held-out) dataset with $N$ and $M$ number of samples. 

% \shafiq{You should try to formally define the learning problem here with the notations.} Addressed

Let's define the training problem first.
%For clean notation, use of indices is minimized whenever possible.
Denote by $\mathcal{D}$ any task dataset consisting of $|\mathcal{D}|$ examples and $\{\mathcal{D}_{t}\}_{t=1}^{T}$ a multitask mixture of $T$ task datasets. Given a multitask mixture of training datasets $\{\mathcal{D}_{t}\}_{t=1}^{T}$, our objective is to train a language model that generalizes better for a test dataset $\mathcal{D}_\text{test}$.

The SPT (semi-parametric prompt tuning) method is depicted in Figure \ref{fig:model}. On a high level, the memory prompt, soft prompt, and input embeddings are concatenated as the input to the language model.
We will describe them as follows.

\subsection{Discrete Prompt}

As shown in Figure \ref{fig:model},
an input example is transformed into a prompted example via a discrete prompt template, then into input embeddings via embedding lookup.
Such input embeddings are the discrete prompt for the given input example.
Likewise, the label (target) of the input example is also transformed 
into a target text sequence via a discrete prompt template (not shown in Figure \ref{fig:model}) for the language model to predict.
Denote by $\mathbf{x}_1, \ldots, \mathbf{x}_q$ the discrete prompt,
and $y_1, \ldots, y_r$ tokens of the target text sequence.

For our training problem, both $\{\mathcal{D}_{t}\}_{t=1}^{T}$ and $\mathcal{D}_{\text{test}}$ are transformed by their respective discrete prompt templates.
Note that a task dataset may have multiple discrete prompt templates.
Specifically,
if there are $p$ discrete prompt templates for the task dataset $\mathcal{D}$, there will be $p \times |\mathcal{D}|$ examples for $\mathcal{D}$.

% \shafiq{The notations need to be more aligned and intuitive} Addressed

\subsection{Memory Prompt}

We will use the discrete prompt to allow the language model
to access an additional memory bank $\mathbf{M}$.
At the very beginning, we initialize $\mathbf{M}$  as a deep copy of the embedding layer of the language model. 
Thus, $\mathbf{M}$ can be considered as a dictionary whose key-value pairs are tokens and their embedding vectors.
This memory bank $\mathbf{M}$ is kept frozen (not trained) all the time.

To retrieve relevant prompt tokens from this memory bank, we 
take the average pooling of $\mathbf{x}_1, \ldots, \mathbf{x}_q$
as the aggregated discrete prompt information.
Then the average pooling result is used to perform the maximum inner product search\footnote{We use the dot-product similarity function.}
to retrieve top-$k_1$ token embeddings from $\mathbf{M}$ 
as the memory prompt: $\mathbf{m}_1, \ldots, \mathbf{m}_{k_1}$.
The memory prompt length $k_1$ is a hyperparameter.

Although the memory prompt is dependent on \emph{every} input example, since the memory bank $\mathbf{M}$ is frozen,
the non-trainable memory prompt is considered as semi-parametric.
We can take this static memory bank as a prior for 
what can be most relevant to the input example 
when constructing a prompted input for a language model.

%\shafiq{You use terms like LM training, MTPL and PEMI, which are quite confusing. I think here you refer to MTPL. You should make it consistent and formal upfront.} -> \maruf{MPTL and PEMI both use the same SPT, So introducing them here will be a bit difficult.}

% \subsection{Objective Function}

% \shafiq{this section needs to be revised as there are quite some issues with the symbols and equations.} -> \maruf{Done}

\subsection{Soft Prompt}

The soft prompt (Figure \ref{fig:model})
is a sequence of trainable token embeddings $\mathbf{s}_1, \ldots, \mathbf{s}_{k_2}$, where the soft prompt length $k_2$ is a hyperparameter.
The soft prompt is initialized as embeddings of (i) downstream task labels and (ii) the most frequent tokens of the tokenizer for the language model,
where embeddings are initialized from a deep copy of the language model's embedding layer.

\subsection{SPT for Multitask Prompted Learning}

In multitask prompted learning,
for any example from the training mixture $\{\mathcal{D}_{t}\}_{t=1}^{T}$ or the test dataset $\mathcal{D}_\text{test}$,
the concatenation of the memory prompt $\mathbf{m}_1, \ldots, \mathbf{m}_{k_1}$, the soft prompt $\mathbf{s}_1, \ldots, \mathbf{s}_{k_2}$, and the discrete prompt $\mathbf{x}_1, \ldots, \mathbf{x}_q$ is inputted to the language model (Figure \ref{fig:model}) for predicting the target $y_1, \ldots, y_r$.

In the following,
we will evaluate SPT under two challenging experimental settings.
The first experimental setting is multitask-inference \emph{full} fine-tuning. Since the full language model will be fine-tuned (e.g., on 31 different tasks from 8 different domains), there is no need to include parameter-efficient soft prompts as the additional trainable parameters ($k_2=0$).
The second experimental setting is multitask-inference \emph{parameter-efficient} fine-tuning.
According to \citet{kaplan_scaling}, the embedding layer is excluded from the scaling law of LLMs, thus we can safely decouple the embedding layer from the language model and keep it in a different device and perform inference at scale. 
Following \citet{SpoT},
during the pre-training stage (e.g., on the GLUE datasets),
the decoupled embedding layer and the soft prompt are trained,
while the rest of the language model is frozen.
During the fine-tuning stage (e.g., on the SuperGLUE datasets),
only the soft prompt is fine tuned.

%For parameter efficient multitask-based downstream fine-tuning, we first decouple the embedding layer from the language model and train only the embedding layer ($\theta = \theta_{E}$) for pre-training, freezing the remaining parameters of the language model. Without considering the embedding layer, our method becomes parameter efficient multitask inference based fine-tuning. 

%Similar to our methods, to enable multitask inference at scale, ATTEMPT \cite{asai2022attempt} requires a small projection layer for calculating the attention vectors. In our work, instead of decoupling the projection layer, we decouple the embedding layer. After pre-training only the embedding layer or multitask mixture, finally,  we only train the soft prompts of the pre-trained model to a target task. 

\section{Experiments}

\subsection{Dataset}
\paragraph{Multitask inference full fine-tuning} We used a sampled version of the P3 dataset that was used to train T0 model \cite{T0}. T0 split of the P3 dataset contains a total of 31 different prompted task datasets from 8 different domains. Instead of taking  $500,000 / \#templates$ per task, we choose a smaller subset of samples (5k samples) from each of the templates. We evaluate zero-shot generalization on 9 datasets in 5 traditional NLP
tasks: (i) natural language inference: CB \cite{Marneffe2019TheCI}, RTE \cite{super_glue} (ii) coreference resolution: WinoGrande XL \cite{ai2:winogrande}, WSC \cite{wino_orig} (iii) word sense disambiguation: WiC \cite{pilehvar-camacho-collados-2019-wic} (iv) sentence completion: COPA \cite{gordon-etal-2012-semeval}, Hellaswag \cite{zellers-etal-2019-hellaswag} (v) question answering: BoolQ \cite{clark-etal-2019-boolq}, MultiRC \cite{khashabi-etal-2018-looking}. Each of the datasets is prompted by multiple discrete prompts from the PromptSource repository \cite{promptsource} \footnote{\url{https://github.com/bigscience-workshop/promptsource}} totaling 87 different task templates. For a fair evaluation, we exclude the task templates that don't generate the full test data.

\paragraph{Multitask inference parameter-efficient fine-tuning} 
\noindent The pre-traing stage in on the GLUE datasets \cite{wang_glue}. Then we perform downstream fine-tuning for 6 different splits of the SuperGLUE benchmark \cite{super_glue}. For fair comparison, we use a single template to transform the training dataset.

For all evaluation datasets, we report evaluation results for all the valid templates available in PromptSource \cite{promptsource}. Following \citet{T0}, we used rank-classification evaluation for models trained on  multitask inference full fine-tuning and generative evaluation \cite{compactor,asai2022attempt} for multitask inference parameter-efficient fine-tuning for a single task adaptation.

\begin{table*}[]
\centering
\scalebox{0.75}{
\begin{tabular}{l|ccc|ccc|ccc}
\toprule
Task  & Small & Small+FT & Small+FT+SPT & Base & Base+FT & Base+FT+SPT & Large & Large+FT & Large+FT+SPT \\
\midrule
\midrule
% \multicolumn{10}{c}{Natural language inference}
\multicolumn{10}{c}{Natural language inference}\\
\midrule
 RTE  & 47.82 & 51.19 & 47.72 & 48.93 & 54.48 & 55.03 & 51.36 & 70.69 & 75.60 \\

 CB  & 40.36 & 41.95 & 43.02 & 36.92 & 41.67 & 53.12 & 35.0 & 71.15 & 71.25 \\
\midrule
\multicolumn{10}{c}{Coreference resolution }\\
\midrule
 Winogrande XL  & 48.64 & 49.93 & 49.75 & 50.06 & 49.67 & 49.9 & 49.74 & 50.95 & 50.38 \\
WSC  & 55.58 & 55.31 & 62.73 & 59.39 & 63.2 & 63.51 & 60.77 & 57.89 & 59.76 \\
\midrule
\multicolumn{10}{c}{Sentence completion}\\
\midrule
 COPA  & 48.68 & 55.76 & 54.44 & 50.29 & 55.37 & 56.15 & 52.89 & 66.6 & 70.21 \\
Hellaswag  & 39.51 & 33.12 & 39.09 & 38.39 & 38.31 & 39.21 & 40.82 & 32.52 & 33.22 \\
\midrule
\multicolumn{10}{c}{Question answering}\\
\midrule
 MultiRC  & 56.26 & 55.81 & 56.06 & 53.93 & 56.65 & 58.2 & 57.17 & 67.8 & 68.29 \\
BoolQ  & 42.56 & 45.83 & 43.45 & 51.59 & 47.74 & 47.93 & 49.9 & 66.9 & 68.41 \\
\midrule
\multicolumn{10}{c}{Word-sense disambiguation}\\
\midrule
  WiC  & 51.12 & 51.18 & 50.20 & 49.86 & 51.75 & 51.89 & 50.56 & 52.03 & 51.80 \\
\midrule
\midrule
 Average  & 47.84 & 48.9 & \textbf{49.61} & 48.82 & 50.98 & \textbf{52.77} & 49.8 & 59.61 & \textbf{60.99} \\
\bottomrule
\end{tabular}%
}
\caption{Multitask-inference full fine-tuning results: zero-shot generalization on 9 heldout datasets under 5 NLP task categories. ``+FT'' means that the full language model is fine-tuned on the T0 split of the P3 dataset consisting of 31 different tasks from 8 different domains. ``+FT+SPT'' means full fine-tuning with SPT.
``Small'', ``Base'' and ``Large'' refer to the vanilla T5 model with different parameter sizes.}
\label{tab:t0_small}
\end{table*}

\subsection{Baselines}

\paragraph{Multitask inference full fine-tuning} We use vanilla T5 models as the weak baseline and full fine-tuning (FT) \cite{T0,FLAN} as the strong baseline for the proposed method. 

\paragraph{Multitask inference parameter-efficient fine-tuning} We use bias fine-tuning \cite{bitfit}, prompt tuning \cite{lester_prompt}, SPoT \cite{SpoT} and ATTEMPT \cite{asai2022attempt} as the baseline multitask inference parameter-efficient fine-tuning methods. We also include the full fine-tuning results for comparison.

\subsection{Experimental Setups}

\noindent  The original prompt tuning paper by \citet{lester_prompt} used T5 \emph{v1.1 LM-adapted} models as the backbone LM. We found that \emph{T5-v1.1 LM-adapted} model is difficult to tune and have convergence issue when used as a backbone LM for prompt tuning. Recent work \cite{compactor,asai2022attempt} also confirms similar findings. Therefore, following \citet{compactor,asai2022attempt}, we use T5 as the backbone model in this work. If a dataset does not include a public test split with annotations, we use the development set as our test set. We used $k_1=20$ and $k_2=0$ for multitask inference full fine-tuning. Following \citet{T0}, in all  experiments, we train the model for a single epoch and perform checkpoint selection by choosing the checkpoint with the highest score on the validation splits of our training datasets. For multitask inference parameter-efficient fine-tuning, the pre-training stage takes a single epoch and the fine-tuning stage takes 10 epochs. For both of the training, we use discretely prompted samples. To compare with current methods \cite{compactor,asai2022attempt}, we use $k_1=10$ and $k_2=90$ ($k_1 + k_2$ is the same as their soft prompt length) for downstream task fine-tuning. We perform five seed experiments and report the average. We strictly maintain the same data order for the same seed in all the experiments. For multitask inference parameter-efficient fine-tuning, we used a learning rate of $0.0001$ for all the experiments. We also tried T0 learning rate $0.001$ but found that $0.0001$ works better with Adam optimizer \footnote{For DeepSpeed-related issues, we were unable to train our language models with the Adafactor optimizer.}. However, for multitask inference parameter-efficient fine-tuning, we ran a hyperparameter search over $\{0.1, 0.3, 0.01, 0.001, 0.001\}$ and found the learning rate $0.3$ gives better convergence. With lower learning rates, we did not observe any convergence of the model.

% Ref is not working
\section{Results}

\begin{table*}[]
\centering
\small
%\resizebox{.9\textwidth}{!}{%
\begin{tabular}{cccccccc}
\toprule
Method & CB & BoolQ & RTE & WiC & WSC & MultiRC & Average \\ 
\midrule
\midrule

Discrete prompt & 27.14 & 28.15 & 39.78 & 38.70 & 32.60 & 51.87 & 36.37 \\ 

\midrule

 Soft prompt & 33.81 & 47.65 & 49.53 & 44.08 & 30.86 & 52.47 & 43.07 \\
\midrule

SPT & 33.45 & 36.13 & 52.02 & 44.06 & 40.77 & 59.51 & \textbf{44.32} \\ 
\bottomrule
\end{tabular}%
%}
\caption{Evaluating the pre-training stage (zero-shot) generalization for multitask-inference parameter-efficient fine-tuning. We pre-train T5-base on the GLUE datasets and evaluate zero-shot performance on the SuperGLUE datasets.}
\label{tab:glue_pretrain}
\end{table*}

%\caption{Multitask-inference full fine-tuning results: zero-shot generalization on 9 heldout datasets under 5 NLP task categories. ``+FT'' means that the full language model is fine-tuned on the T0 split of the P3 dataset consisting of 31 different tasks from 8 different domains. ``+FT+SPT'' means full fine-tuning with SPT. ``Small'', ``Base'' and ``Large'' refer to the vanilla T5 with different parameter sizes.}
\begin{table*}
\small
\centering
%\resizebox{\textwidth}{!}{%
\begin{tabular}{cccccccc}
\toprule
Method                                                                                                                & CB    & BoolQ & RTE   & WiC   & WSC & \multicolumn{1}{l}{MultiRc} & Average   \\
\midrule
\midrule
%\multicolumn{8}{c}{High-performance methods}                                    
                                                                                                            
%\midrule
Full fine-tuning                                                                                                        & 85.71 & 81.10  & 71.94 & 70.22 & 59.61     & 72.77                       & 73.56 \\
% Adapter              & 85.71 & 82.45 & 71.94 & 67.08 & 67.30      & 75.87                       & 75.06 \\
\midrule
\midrule
%\multicolumn{8}{c}{Multitask-inference parameter-efficient fine-tuning methods}                                                                                                                             %                                  \\
%\midrule

\citet{bitfit}                                                                                                                       & 67.63 & 79.57 & 67.63 & 69.59 & 59.60      & 74.51                       & 69.76 \\
\citet{lester_prompt}                                                                                                           & 67.86 & 61.71 & 54.68 & 48.90  & 51.92     & 58.73                       & 57.30  \\
\citet{SpoT}                                                                                                                 & 71.43 & 71.68 & 71.94 & 48.90  & 53.84     & 74.21                       & 65.33 \\
 \citet{asai2022attempt}                                                                                                               & 78.57 & 77.06 & 73.38 & 66.77 & 53.84     & 74.39                       & 70.67 \\
SPT            & \textbf{85.35} & \textbf{80.55} & \textbf{79.78} & \textbf{61.91} & \textbf{65.19}     & \textbf{75.18}                       & \textbf{74.66}\\
\bottomrule
\end{tabular}%
%}
\caption{Multitask-inference parameter-efficient fine-tuning results on the SuperGLUE datasets after T5-base is pre-trained on the GLUE datasets.}
\label{tab:super_glue}
\end{table*}

 For multitask-inference full fine-tuning, we compare our proposed method with full fine-tuning and vanilla T5 model. Following \citet{T0,FLAN}, we see improved performance gain with multitasking pre-training. However, our proposed SPT method outperforms pure multitask fine-tuning \cite{T0} by a large margin. For ``small'', ``base'' and ``large'' models we see an absolute average improvement of +0.71, +1.79 and +1.38 over 9 different evaluation datasets on 87 different templates. Table \ref{tab:t0_small} shows the average results on all the tasks. The full version of the results for each of the prompt templates is added in the Appendix (Table \ref{tab:t0}).

Table \ref{tab:glue_pretrain} evaluates the pre-training stage (zero-shot) generalization for multitask-inference parameter-efficient
fine-tuning. For this experiment, We train each task on a single template from the GLUE dataset and evaluate it on the 6 tasks from SuperGLUE benchmark. In this set of experiments at first, as a discrete prompt baseline, we train only the embedding layer of the language model and achieve around $36.37$ average scores. In addition to the embedding layer, we also add the soft prompt \cite{lester_prompt} with the embedding layer to train the model and see improved performance. Finally, we compare it with our proposed SPT method and see an overall average improvement of $+1.25$. 

Table \ref{tab:super_glue} shows the result of multitask-inference parameter-efficient fine-tuning on the six tasks of the SuperGLUE benchmark. In this stage, we take the previously pre-trained model (from Table \ref{tab:glue_pretrain}) on the GLUE tasks and train only the soft prompts of the model for a single task (from SuperGLUE tasks) fine-tuning. Overall in all the datasets, we see an absolute average improvement of $+3.99$ over the previous multitask-inference parameter-efficient fine-tuning method.

\section{Analysis}

\begin{figure*}%
    \centering
    %{\includegraphics[scale=.3]{figs/scaling-law-model.png}}%
    {\includegraphics[scale=.45]{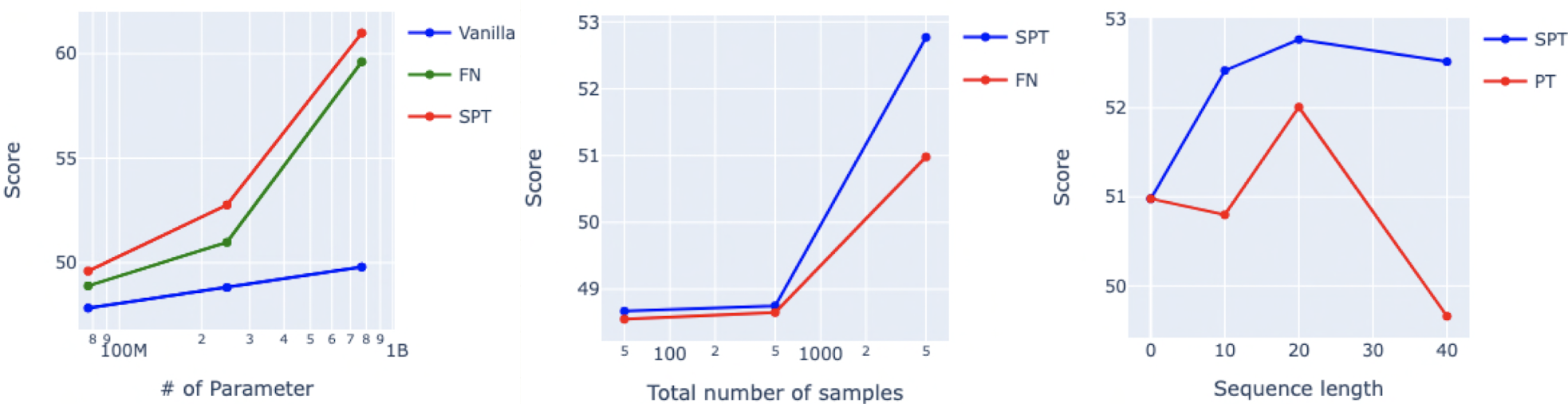}}%
    
    \caption{Different scaling laws for our proposed SPT method. For both (a) and (b), the logarithmic (10) scale is applied to the $x$-axis. In (a), each of the 3 points in the line indicates the score of T5-small/base/large, respectively. In (b) we observe that with same amount of data, SPT performs better than full fine-tuning. In (c) we see that as we increase the length of the memory prompt, SPT performs reasonably well compared to prompt-tuning.}%
    \label{fig:scaling_model}%
\end{figure*}

\begin{figure*}%
    \centering
    %{\includegraphics[scale=.3]{figs/scaling-law-model.png}}%
    {\includegraphics[scale=.45]{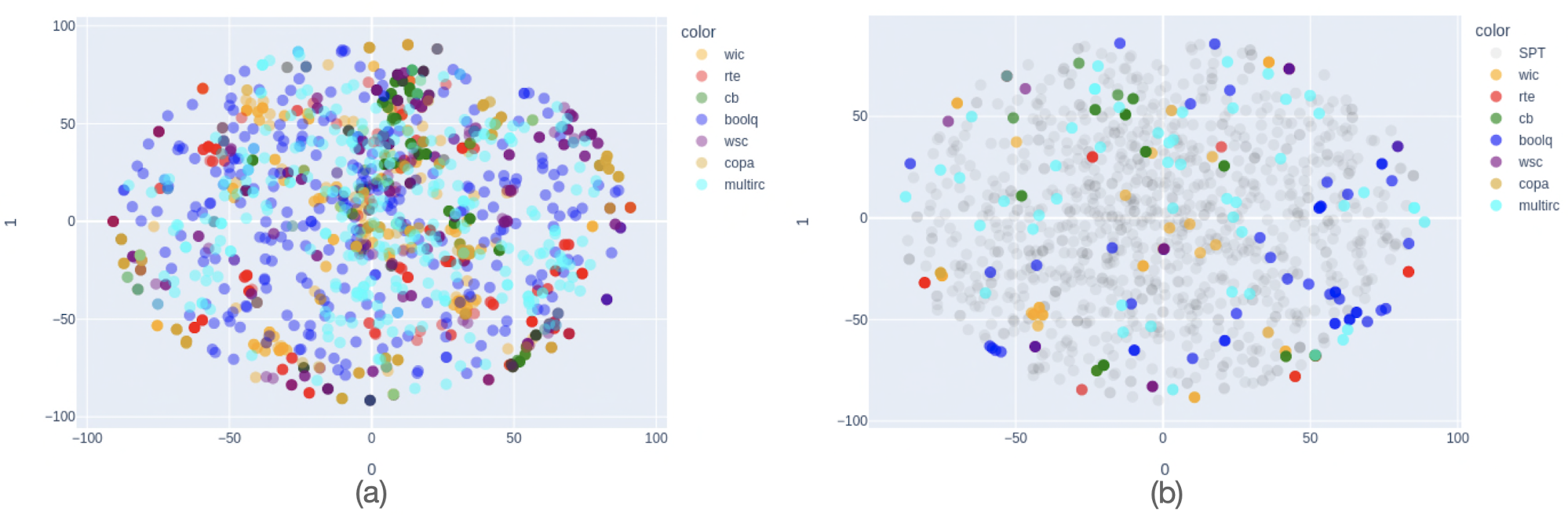}}%
    
    \caption{The t-SNE plot for memory prompt distribution of SuperGLUE tasks. (a) shows the full distribution of all the memory prompts, and (b) includes additional colored markers indicating instances where a model without a memory prompt predicts the wrong class, but a model with a memory prompt predicts the correct class. For both (a) and (b), there is no task-specific cluster which indicates a successful multitask prompting.}%
    \label{fig:tsne}%
\end{figure*}

\begin{figure*}%
    % https://drive.google.com/file/d/1AJrFPUS08cUXOZymzMlX6Xz6KSEU2H6r/view?usp=sharing
    \centering
    {\includegraphics[scale=.53]{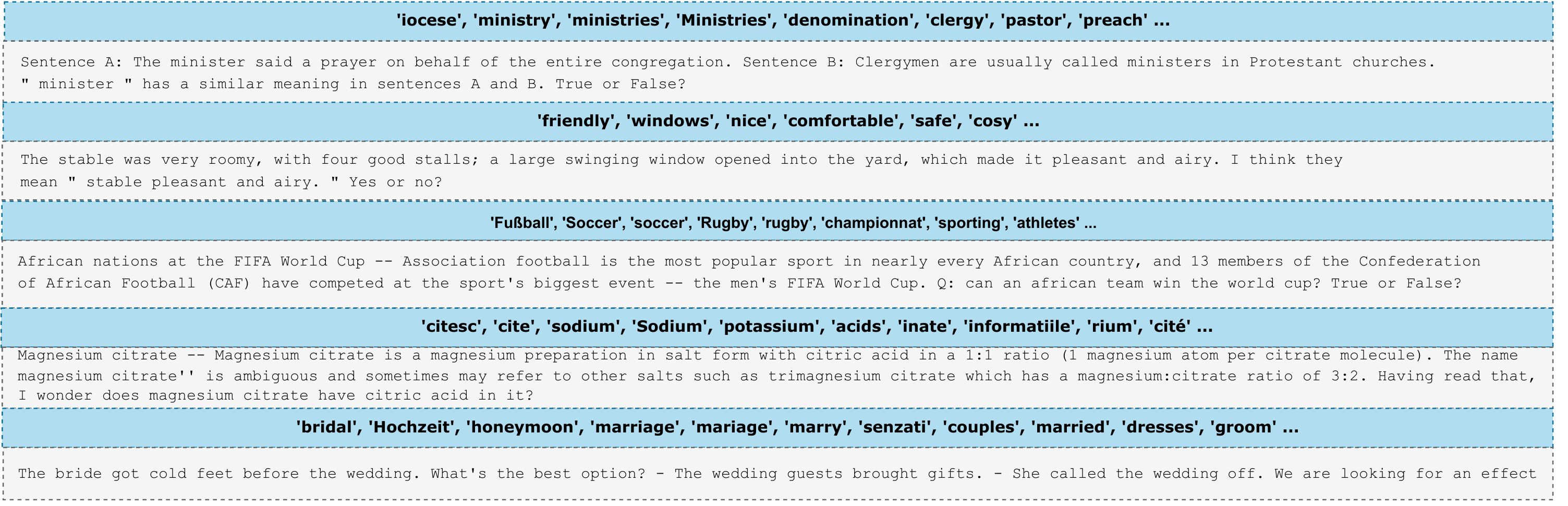}}%
    \caption{Examples of memory prompts (in the blue box).  For all of the samples (in the gray box), full model fine-tuning (FT) fails, and the SPT successfully predicts the true class.}%
    \label{fig:spt-prompts}%
\end{figure*}

\paragraph{Scaling law} Figure \ref{fig:scaling_model} shows the different scaling properties of SPT for multitask-inference full fine-tuning. First, we show the scaling law for different backbone LMs in Figure \ref{fig:scaling_model}(a). To calculate the score for each of the models, we aggregate the average rank classification score for all the prompts from our selected evaluation tasks. As the model grows bigger, we see overall  performance improvement for T5-small/base/large. 

In Figure \ref{fig:scaling_model}(b) we also explore the influence of data during pre-training. To test how data diversity contributes to model performance, we randomly sample $50$, $500$ and $5000$ examples from each of the templates of the pre-training tasks. In Figure \ref{fig:scaling_model}(b), although SPT performs well compared to full fine-tuning, we have come to the conclusion that in the absence of an adequate number of samples per template, it is not possible to improve model performance using either fine-tuning (FT) or SPT.

Finally, we explore the effect of sequence length for multitask-inference full fine-tuning. In Figure \ref{fig:scaling_model}(c), we observe that compared to soft prompts, memory prompts perform reasonably well for all the sequence length. We also observe diminishing return for memory prompt and huge negative return for soft-prompt as the prompt length grows larger (40). This also indicates that memory prompts are more robust.

\paragraph{Memory prompt distribution} In Figure \ref{fig:tsne} we analyze the memory prompt distributions. In Figure \ref{fig:tsne}(a), we plot the memory prompt distribution of SuperGLUE tasks. In Figure \ref{fig:tsne}(b), we compare the distribution of memory prompts with the memory prompts of the samples where the model that receives a memory prompt correctly predicts the class, but the model without a memory prompt does not. In both Figure \ref{fig:tsne}(a) and Figure \ref{fig:tsne}(b), we observe that there is no task-specific cluster, and the distribution is wide and spread. Since in multitask inference full fine-tuning, we ground all the task inputs into a set of prompted examples (instructions), it indicates a successful multitask prompting.

\paragraph{Memory prompt tokens} After training the multitask-inference full fine-tuning model, we analyze the retrieved memory prompt tokens. Figure \ref{fig:spt-prompts} shows a few examples with unique memory prompt tokens that are not part of the input tokens. For all the samples in Figure \ref{fig:spt-prompts}, full model fine-tuning is not able to find correct predictions, but the model that uses memory prompt successfully performs rank classification prediction. As we use dot product similarity during training, in most cases, we find that the memory prompts are tokens that are contextually comparable to the inputs or they are synonyms of the inputs. However, we also observe token repetitiveness with different capitalization as well as punctuation in the memory prompt.

\section{Releted Work}

%There have been the following two lines of studies that are relevant to our work.

\paragraph{Multitask prompted learning} Multitask learning has been shown to improve the performance of NLP models \cite{collobert}. For explicit multitask learning, augmenting all the samples during training may or may not induce noise due to different output distributions in a traditional full-model fine-tuning setup \cite{weller-etal-2022-use,bari2021}. For implicit multitask learning \citet{radford2019language} showed that the language model begins to learn many downstream tasks without any explicit supervision during pre-training. At scale, large language models \cite{gpt-3,turing-nlg,palm,bloom} can also perform few-shot in-context evaluation, which makes it an effective multi-task model. 

Finally, \citet{T0,FLAN,bloomz,flan_t5} showed that these implicitly learned language models can be further improved by explicitly fine-tuning discretely promoted human instruction \cite{promptsource,super_natural} in a multitask fashion.

\paragraph{Parameter-efficient fine-tuning}

The increasing size of large language models (LLMs) \cite{palm,turing-nlg,switch_trainsformer} can make it harder to fine-tune them to a new target task. Compared to smaller models, fine-tuning these big LMs frequently results in a performance reduction, especially in low-resource scenarios \cite{t-few,bari2021nearest}. Moreover, pre-trained LMs may be sensitive to the initialization of the fine-tuning procedure, which might further affect the model's performance on the new job. Working with big pre-trained LMs necessitates careful consideration of the fine-tuning technique. Parameter-efficient fine-tuning, in contrast to full fine-tuning, only modifies a handful of a number of parameters \cite{visual_adapters,Houlsby_peft,adapter_nmt,guo-etal-2021-parameter}. Recent work \cite{lester_prompt,he-etal-2021-effectiveness} has also discovered that parameter-efficient fine-tuning can keep the generic pre-trained distribution since it freezes most of the parameters and avoids the issue of forgetting or overwriting distribution. Parameter-efficient fine-tuning is applied to the language models in many different ways, notably (i) adding low-rank updates \cite{Lora,t-few} and hyper-complex layers \cite{aston_quaternions,compactor,IPDG}; (ii) training a small subset of parameters from the language model \cite{ruckle-etal-2021-adapterdrop,pfeiffer-etal-2020-mad,pfeiffer-etal-2021-adapterfusion} or directly modifying few parameters inside the transformer block \cite{ben-zaken-etal-2022-bitfit}; (iii) prepending soft prompt with the input embeddings \cite{lester_prompt,SpoT,asai2022attempt} or activation layers \cite{li2021prefixtuning} of the Transformer encoder. 

% TODO: Cite IPDG
\section{Conclusion}
We presented SPT (semi-parametric prompt tuning), a new prompt-tuning approach for multitask prompted learning. Our proposed method initializes a memory bank from the input distribution and retrieve an input-specific memory prompt by performing maximum inner product search. Experiments on a wide range of evaluation datasets demonstrated effectiveness of SPT in two challenging experimental settings: multitask-inference full fine-tuning and multitask-inference parameter-efficient fine-tuning.
\section{Limitations}
Our proposed method still uses human-written prompts to augment NLP datasets and perform a query on the memory to retrieve semi-parametric prompts. Thus, the retrieved prompts are subject to annotator bias and diversity. %In addition to that, we haven't shown the scaling law with a large-scale language model. 

\bibliography{main,anthology}

\paragraph{Dataset Mixtures}
The dataset description of each of the multitask mixtures is given in Table \ref{tab:datasets}.
\begin{table*}[]
\resizebox{\textwidth}{!}{%
\begin{tabular}{llrccc}
\toprule
\multicolumn{6}{c}{Training Datasets}  \\
\midrule
\midrule
No & Task & Task Type & \# of Train Samples & \# of Dev Samples & \# of Templates \\ 
\midrule
\multicolumn{6}{c}{PEMI Mixture}  \\
\midrule
1  & Cola & Grammatical Acceptability & 8551                & 1043              & 1               \\
2  & SST2 & Sentiment Analysis        & 67349               & 872               & 1              \\
3  & MRPC & Paraphrase Identification        & 3668               & 408               & 1              \\
4  & QQP & Paraphrase Identification        & 363846         &  40430              & 1              \\
5  & MNLI & Natural Language Inference        & 392702               & 19647               & 1              \\
6  & QNLI & Natural Language Inference        & 104743               & 5463               & 1              \\
7  & RTE & Natural Language Inference        & 2490               & 277               & 1              \\
8  & WNLI & Natural Language Inference         & 635               & 71               & 1              \\
\midrule
\multicolumn{6}{c}{MTPL Mixture (A subset of promptsource \cite{promptsource})}  \\
\midrule
1  & Adversarial QA & Extractive QA & 5000                & 25              & 15               \\
2  & Quoref & Extractive QA        & 5000               & 25               & 11              \\
3  & ROPES & Extractive QA        & 5000               & 25               & 12              \\
4  & DuoRC & Extractive QA        & 5000         &  25              & 18              \\

5  & DREAM & Multiple-Choice QA        & 5000               & 25               & 5             \\
6  & QuAIL & Multiple-Choice QA        & 5000               & 25               & 13              \\
7  & QuaRTz & Multiple-Choice QA        & 5000               & 25               & 13              \\
8  & Social IQA & Multiple-Choice QA        & 5000               & 25               & 6              \\
9  & WiQA & Multiple-Choice QA        & 5000               & 25               & 8              \\
10 & Commonsense QA &   Multiple-Choice QA & 5000 & 25 & 11 \\
11  & Cosmos QA & Multiple-Choice QA        & 5000               & 25               & 13              \\
12  & QASC & Multiple-Choice QA        & 5000               & 25               & 8              \\
13  & SciQ & Multiple-Choice QA        & 5000               & 25               & 5              \\
14  & Wiki Hop & Multiple-Choice QA        & 5000               & 25               & 9              \\
           
15 & AG News & Topic Classification & 5000 & 25 & 7\\
16 & DBPedia & Topic Classification & 5000 & 25 & 4\\
17 & TREC & Topic Classification & 5000 & 25 & 18\\

18 & App Reviews & Sentiment & 5000 & 25 & 4\\
19 & IMDB & Sentiment & 5000 & 25 & 11\\
20 & Rotten Tomatoes & Sentiment & 5000 & 25 & 10\\
21 & Yelp & Sentiment & 5000 & 25 & 7\\

22 & Hotpot QA & Close-Bool QA & 5000 & 25 & 5\\
23 & Wiki QA & Close-Bool QA & 5000 & 25 & 11\\

24 & Common Gen & Structure-To-Text & 5000 & 25 & 9\\
25 & Wiki Bio & Structure-To-Text & 5000 & 25 & 5\\

26 & CNN Daily Mail QA & Summarization & 5000 & 25 & 5\\
27 & Gigaword QA & Summarization & 5000 & 25 & 9\\
28 & MultiNews & Summarization & 5000 & 25 & 7\\
29 & SamSum & Summarization & 5000 & 25 & 7\\
30 & XSum & Summarization & 5000 & 25 & 10\\

31 & MRPC & Paraphrase Identification & 5000 & 25 & 7\\
32 & PAWS & Paraphrase Identification & 5000 & 25 & 12\\
33 & QQP & Paraphrase Identification & 5000 & 25 & 6\\
\midrule
\multicolumn{6}{c}{Evaluation Datasets}  \\
\midrule
\midrule
1 & BoolQ & Yes/NO QA & x & 3270 & 10\\
2 & CB & Natural Language
Inference & x & 56 & 15\\
3 & COPA & Sentence Completion & x & 100 & 8\\
4 & Hellaswag & Sentence Completion & x & 25 & 7\\
5 & MultiRC & Paraphrase Identification & x & 4848 & 10\\
6 & RTE & Natural Language Inference & x & 277 & 10\\
7 & WiC & Word Sense Disambiguation & x & 638 & 10\\
8 & Winogrande XL & Coreference Resolution & x & 1267 & 7\\
9 & WSC & Coreference Resolution & x & 104 & 10\\
\bottomrule

\end{tabular}%
}
\caption{Dataset statistics of our multitask mixture. We mix all the tasks on 1:1 basis. For tasks with more than a single prompt template, we augment all the samples by each of the prompt template $5000 \time \# \ of \ template $ samples per task.}
\label{tab:datasets}
\end{table*}

\paragraph{Evaluation Results} Evaluation result for each of the specific prompted dataset is added in the Table \ref{tab:t0}.
\begin{table*}[]
\centering
\scalebox{0.48}{
\begin{tabular}{ccccccccccccccccccc}
\toprule
Task & Prompt  & Small & Small+FN & Small+FN+SPT & Base & Base+FN & Base+FN+SPT & Large & Large+FN & Large+FN+SPT \\
\midrule
WiC & GPT-3-prompt-with-label
& 50.62 & 52.47 & 49.74 & 49.48 & 50.31 & 50.31 & 53.59 & 53.28 & 52.66 \\
WiC & polysemous
& 50.31 & 52.6 & 49.74 & 49.09 & 50.16 & 50.16 & 50.16 & 54.37 & 52.81 \\
WiC & question-context-meaning-with-label
& 50.31 & 46.74 & 50.13 & 49.61 & 50.62 & 51.72 & 50.16 & 54.53 & 54.53 \\
WiC & similar-sense
& 53.44 & 50.65 & 50.0 & 50.13 & 51.72 & 52.03 & 50.78 & 48.91 & 48.44 \\
WiC & grammar homework
& 50.16 & 51.56 & 50.91 & 49.87 & 52.03 & 51.41 & 50.16 & 50.31 & 50.16 \\
WiC & question-context
& 50.62 & 54.04 & 50.52 & 50.65 & 50.94 & 52.66 & 49.38 & 50.16 & 50.47 \\
WiC & GPT-3-prompt
& 50.16 & 52.73 & 50.0 & 50.26 & 55.47 & 55.47 & 50.31 & 53.44 & 53.91 \\
WiC & question-context-meaning
& 52.81 & 48.7 & 51.17 & 50.0 & 52.5 & 51.56 & 50.78 & 50.16 & 50.16 \\
WiC & affirmation true or false
& 52.03 & 49.22 & 50.0 & 49.74 & 52.19 & 53.28 & 50.16 & 50.31 & 50.62 \\
WiC & same sense
& 50.78 & 53.12 & 49.74 & 49.74 & 51.56 & 50.31 & 50.16 & 54.84 & 54.22 \\
\midrule
 & AVG WiC  & 51.12 & 51.18 & 50.2 & 49.86 & 51.75 & 51.89 & 50.56 & 52.03 & 51.8 \\
\midrule
RTE & can we infer
& 47.14 & 47.27 & 46.88 & 46.88 & 46.09 & 47.92 & 47.5 & 72.81 & 78.39 \\
RTE & should assume
& 47.5 & 52.54 & 48.63 & 47.27 & 55.21 & 57.55 & 47.5 & 71.88 & 75.0 \\
RTE & does this imply
& 47.86 & 48.44 & 46.88 & 51.95 & 52.08 & 51.82 & 62.86 & 74.38 & 77.08 \\
RTE & based on the previous passage
& 47.14 & 53.32 & 46.88 & 47.27 & 62.5 & 58.59 & 47.5 & 71.56 & 75.78 \\
RTE & must be true
& 47.14 & 50.0 & 46.88 & 46.88 & 60.42 & 61.72 & 47.5 & 71.25 & 74.48 \\
RTE & does it follow that
& 47.14 & 50.78 & 46.88 & 46.68 & 48.18 & 49.22 & 47.5 & 71.88 & 77.34 \\
RTE & GPT-3 style
& 52.86 & 55.66 & 52.93 & 60.94 & 59.11 & 59.64 & 70.71 & 62.19 & 73.44 \\
RTE & justified in saying
& 47.14 & 50.2 & 46.88 & 46.88 & 46.35 & 47.66 & 47.5 & 68.44 & 75.0 \\
RTE & guaranteed true
& 47.14 & 52.93 & 46.88 & 47.27 & 58.33 & 58.33 & 47.5 & 71.56 & 73.7 \\
RTE & MNLI crowdsource
& 47.14 & 50.78 & 47.46 & 47.27 & 56.51 & 57.81 & 47.5 & 70.94 & 75.78 \\
\midrule
 & AVG RTE  & 47.82 & 51.19 & 47.72 & 48.93 & 54.48 & 55.03 & 51.36 & 70.69 & 75.6 \\
\midrule
CB & does this imply
& 50.0 & 43.36 & 40.23 & 42.19 & 32.81 & 57.03 & 50.0 & 81.25 & 81.25 \\
CB & should assume
& 50.0 & 40.23 & 40.23 & 50.39 & 35.16 & 61.72 & 50.0 & 78.12 & 76.56 \\
CB & can we infer
& 50.0 & 40.23 & 40.23 & 41.07 & 23.44 & 55.47 & 50.0 & 81.25 & 79.69 \\
CB & claim true false inconclusive
& 17.86 & 50.39 & 50.39 & 16.41 & 60.16 & 65.62 & 10.71 & 68.75 & 85.16 \\
CB & take the following as truth
& 10.71 & 50.39 & 50.39 & 10.94 & 43.75 & 53.12 & 8.93 & 60.94 & 66.41 \\
CB & must be true
& 50.0 & 40.23 & 40.23 & 42.97 & 50.78 & 61.72 & 50.0 & 76.56 & 76.56 \\
CB & guaranteed possible impossible
& 33.93 & 48.44 & 50.39 & 36.33 & 45.31 & 50.78 & 8.93 & 57.81 & 52.34 \\
CB & based on the previous passage
& 50.0 & 40.23 & 40.23 & 46.48 & 46.09 & 52.34 & 50.0 & 82.81 & 81.25 \\
CB & always sometimes never
& 39.29 & 40.23 & 42.19 & 32.42 & 37.5 & 39.06 & 8.93 & 50.0 & 50.0 \\
CB & does it follow that
& 48.21 & 40.23 & 38.28 & 46.48 & 39.84 & 55.47 & 50.0 & 76.56 & 80.47 \\
CB & MNLI crowdsource
& 37.5 & 46.48 & 44.53 & 9.38 & 48.44 & 39.84 & 10.71 & 79.69 & 72.66 \\
CB & guaranteed true
& 46.43 & 40.23 & 40.23 & 52.34 & 46.88 & 60.16 & 50.0 & 78.12 & 77.34 \\
CB & justified in saying
& 50.0 & 40.23 & 40.23 & 48.21 & 28.91 & 57.81 & 50.0 & 84.38 & 83.59 \\
CB & consider always sometimes never
& 37.5 & 53.52 & 52.73 & 27.73 & 37.5 & 39.06 & 10.71 & 59.38 & 47.66 \\
CB & GPT-3 style
& 33.93 & 14.84 & 34.77 & 50.39 & 48.44 & 47.66 & 66.07 & 51.56 & 57.81 \\
\midrule
 & AVG CB  & 40.36 & 41.95 & 43.02 & 36.92 & 41.67 & 53.12 & 35.0 & 71.15 & 71.25 \\
\midrule
Hellaswag & Appropriate continuation - Yes or No
& 74.74 & 53.62 & 71.57 & 71.84 & 74.2 & 74.28 & 75.13 & 50.06 & 50.4 \\
Hellaswag & if begins how continues
& 25.14 & 25.35 & 25.2 & 25.1 & 25.24 & 25.36 & 24.99 & 28.76 & 28.26 \\
Hellaswag & complete first then
& 24.76 & 24.77 & 24.96 & 25.3 & 25.88 & 25.43 & 25.69 & 26.27 & 27.25 \\
Hellaswag & Predict ending with hint
& 25.21 & 25.91 & 25.47 & 25.34 & 26.03 & 25.52 & 26.03 & 27.19 & 27.88 \\
Hellaswag & Randomized prompts template
& 25.4 & 25.34 & 25.02 & 24.95 & 25.25 & 24.4 & 25.44 & 26.21 & 27.06 \\
Hellaswag & Reversed appropriate continuation - Yes or No
& 74.79 & 49.57 & 74.49 & 68.36 & 62.36 & 70.36 & 75.22 & 36.42 & 39.05 \\
Hellaswag & Open-ended completion
& 26.55 & 27.29 & 26.95 & 27.81 & 29.21 & 29.09 & 33.25 & 32.72 & 32.62 \\
\midrule
 & AVG Hellaswag  & 39.51 & 33.12 & 39.09 & 38.39 & 38.31 & 39.21 & 40.82 & 32.52 & 33.22 \\
\midrule
BoolQ & could you tell me…
& 37.9 & 38.16 & 37.41 & 38.16 & 42.58 & 44.29 & 37.84 & 71.63 & 68.72 \\
BoolQ & yes no question
& 39.7 & 37.2 & 37.44 & 63.73 & 38.16 & 37.98 & 70.02 & 74.07 & 76.17 \\
BoolQ & valid binary
& 60.97 & 61.03 & 56.76 & 56.85 & 53.88 & 55.11 & 38.42 & 50.9 & 61.75 \\
BoolQ & exercise
& 39.7 & 48.32 & 46.0 & 69.23 & 61.96 & 61.99 & 70.39 & 54.36 & 57.39 \\
BoolQ & based on the following passage
& 38.45 & 39.39 & 38.19 & 37.89 & 46.06 & 45.16 & 38.42 & 72.39 & 70.43 \\
BoolQ & based on the previous passage
& 37.99 & 38.61 & 37.98 & 37.95 & 47.09 & 46.33 & 37.87 & 73.95 & 71.69 \\
BoolQ & I wonder…
& 37.9 & 38.82 & 37.62 & 38.13 & 44.68 & 47.06 & 37.9 & 71.69 & 69.89 \\
BoolQ & GPT-3 Style
& 46.27 & 40.02 & 40.5 & 63.94 & 39.48 & 40.59 & 46.15 & 74.97 & 71.39 \\
BoolQ & after reading
& 48.93 & 60.01 & 54.57 & 63.61 & 61.3 & 58.23 & 63.17 & 51.68 & 62.8 \\
BoolQ & exam
& 37.78 & 56.79 & 47.99 & 46.45 & 42.19 & 42.58 & 58.83 & 73.35 & 73.83 \\
\midrule
 & AVG BoolQ  & 42.56 & 45.83 & 43.45 & 51.59 & 47.74 & 47.93 & 49.9 & 66.9 & 68.41 \\
\midrule
WSC & does the pronoun refer to
& 66.35 & 46.09 & 63.28 & 63.46 & 64.06 & 64.06 & 63.46 & 47.66 & 52.34 \\
WSC & in other words
& 39.42 & 63.28 & 63.28 & 63.28 & 62.5 & 63.28 & 63.46 & 64.84 & 60.94 \\
WSC & does p stand for
& 61.54 & 51.95 & 63.28 & 61.72 & 64.06 & 64.06 & 63.46 & 47.66 & 57.81 \\
WSC & the pronoun refers to
& 38.46 & 63.28 & 63.28 & 63.28 & 61.72 & 64.06 & 63.46 & 64.84 & 64.06 \\
WSC & p is are r
& 57.69 & 63.28 & 63.28 & 36.72 & 64.06 & 64.06 & 36.54 & 64.84 & 65.62 \\
WSC & by p they mean
& 63.46 & 41.8 & 63.28 & 63.28 & 64.06 & 63.28 & 63.46 & 58.59 & 58.59 \\
WSC & Who or what is are
& 51.92 & 63.28 & 62.5 & 53.52 & 64.06 & 64.84 & 63.46 & 64.84 & 64.06 \\
WSC & GPT-3 Style
& 49.04 & 58.98 & 58.59 & 63.28 & 61.72 & 61.72 & 63.46 & 54.69 & 60.16 \\
WSC & replaced with
& 63.46 & 57.42 & 63.28 & 63.28 & 64.06 & 62.5 & 63.46 & 63.28 & 64.06 \\
WSC & I think they mean
& 64.42 & 43.75 & 63.28 & 62.11 & 61.72 & 63.28 & 63.46 & 47.66 & 50.0 \\
\midrule
 & AVG WSC  & 55.58 & 55.31 & 62.73 & 59.39 & 63.2 & 63.51 & 60.77 & 57.89 & 59.76 \\
\midrule
Winogrande XL & True or False
& 51.1 & 49.53 & 49.69 & 50.39 & 50.7 & 51.88 & 50.39 & 50.39 & 50.39 \\
Winogrande XL & Replace
& 49.53 & 50.08 & 50.62 & 49.69 & 49.53 & 50.16 & 50.0 & 52.03 & 50.78 \\
Winogrande XL & fill in the blank
& 46.48 & 50.16 & 50.0 & 49.69 & 49.92 & 50.47 & 49.37 & 51.72 & 50.16 \\
Winogrande XL & stand for
& 49.06 & 49.53 & 49.22 & 50.23 & 48.59 & 49.14 & 50.0 & 49.84 & 50.08 \\
Winogrande XL & underscore refer to
& 48.67 & 50.39 & 49.53 & 50.39 & 49.61 & 48.36 & 50.0 & 50.62 & 50.39 \\
Winogrande XL & jsonl
& 46.48 & 50.16 & 50.0 & 49.69 & 49.92 & 50.47 & 49.45 & 51.72 & 50.16 \\
Winogrande XL & does underscore refer to
& 49.14 & 49.69 & 49.22 & 50.31 & 49.45 & 48.83 & 48.98 & 50.31 & 50.7 \\
\midrule
 & AVG Winogrande XL  & 48.64 & 49.93 & 49.75 & 50.06 & 49.67 & 49.9 & 49.74 & 50.95 & 50.38 \\
\midrule
COPA & exercise
& 48.08 & 55.47 & 58.59 & 50.39 & 50.78 & 55.47 & 49.04 & 66.41 & 69.53 \\
COPA & more likely
& 48.08 & 55.47 & 56.25 & 50.0 & 55.47 & 57.81 & 56.73 & 64.84 & 67.97 \\
COPA & plausible alternatives
& 50.0 & 52.73 & 51.56 & 48.05 & 56.25 & 63.28 & 56.73 & 71.88 & 74.22 \\
COPA & best option
& 48.08 & 55.47 & 52.73 & 50.0 & 57.03 & 50.78 & 53.85 & 64.06 & 70.31 \\
COPA & choose
& 49.04 & 56.64 & 51.95 & 48.83 & 59.38 & 57.03 & 50.96 & 69.53 & 71.09 \\
COPA & i am hesitating
& 47.12 & 54.3 & 52.73 & 52.73 & 55.47 & 57.81 & 49.04 & 70.31 & 75.78 \\
COPA & cause effect
& 49.04 & 57.42 & 54.69 & 45.31 & 55.47 & 55.47 & 53.85 & 67.19 & 67.97 \\
COPA & C1 or C2? premise, so because…
& 50.0 & 58.59 & 57.03 & 57.03 & 53.12 & 51.56 & 52.88 & 58.59 & 64.84 \\
\midrule
 & AVG COPA  & 48.68 & 55.76 & 54.44 & 50.29 & 55.37 & 56.15 & 52.89 & 66.6 & 70.21 \\
\midrule
MultiRC & confirm
& 55.34 & 56.21 & 56.25 & 54.38 & 57.2 & 58.88 & 57.92 & 66.47 & 66.9 \\
MultiRC & Would it be good to answer…
& 56.06 & 55.06 & 55.16 & 53.64 & 56.93 & 57.46 & 57.08 & 69.26 & 70.27 \\
MultiRC & paragraph… question… is it… ?
& 55.34 & 56.21 & 56.33 & 51.07 & 58.08 & 58.92 & 55.98 & 67.15 & 63.92 \\
MultiRC & found this answer
& 55.67 & 56.15 & 56.31 & 53.99 & 59.0 & 60.42 & 57.41 & 69.22 & 70.39 \\
MultiRC & correct
& 58.07 & 57.57 & 56.78 & 59.5 & 58.82 & 59.56 & 57.63 & 70.87 & 71.48 \\
MultiRC & I was going to say…
& 56.6 & 53.89 & 55.16 & 54.61 & 53.27 & 55.47 & 56.93 & 63.18 & 63.03 \\
MultiRC & grading
& 55.88 & 55.63 & 55.47 & 51.52 & 57.3 & 59.66 & 55.59 & 69.47 & 70.66 \\
MultiRC & is the correct answer…
& 55.92 & 57.09 & 56.64 & 50.64 & 57.52 & 58.96 & 59.16 & 69.55 & 70.02 \\
MultiRC & is… a correct answer?
& 56.72 & 55.84 & 56.25 & 53.91 & 51.52 & 54.89 & 56.93 & 66.78 & 69.04 \\
MultiRC & decide valid
& 56.99 & 54.46 & 56.29 & 56.0 & 56.89 & 57.73 & 57.12 & 66.08 & 67.15 \\
\midrule
 & AVG MultiRC  & 56.26 & 55.81 & 56.06 & 53.93 & 56.65 & 58.2 & 57.17 & 67.8 & 68.29 \\
\midrule
 & AVG.  & 47.84 & 48.9 & \textbf{49.61} & 48.82 & 50.98 &  \textbf{52.77} & 49.8 & 59.61 &  \textbf{60.99} \\
\midrule
\bottomrule
\end{tabular}%
}
\caption{Full version of Table \ref{tab:t0_small}.}
\label{tab:t0}
\end{table*}

\end{document}